\documentclass[runningheads]{llncs}
\usepackage{graphicx}

\usepackage{tikz}
\usepackage{comment}
\usepackage{amsmath,amssymb} 
\usepackage{color}
\usepackage{comment}
\usepackage{color}
\usepackage{siunitx}
\usepackage{booktabs}
\usepackage{bm}
\usepackage{mathtools}
\usepackage{xcolor}
\usepackage{floatrow}
\usepackage{multirow}
\usepackage{wrapfig,lipsum,booktabs}
\usepackage{caption}
\usepackage{subcaption}
 \usepackage{sidecap}
\usepackage{xspace}
\usepackage{textcomp}
\usepackage{relsize}
\usepackage{amssymb}
\usepackage{pifont}

\usepackage[accsupp]{axessibility}  

\definecolor{citecolor}{RGB}{59,155,85}
\usepackage[pagebackref,breaklinks,colorlinks]{hyperref}

\usepackage{xspace}
\usepackage[export]{adjustbox}
\usepackage[abs]{overpic}
\usetikzlibrary{arrows.meta}

\usepackage{bold-extra}
\usepackage{cite}

\newcommand*{\eg}{\emph{e.g.}\@\xspace}
\newcommand*{\ie}{\emph{i.e.}\@\xspace}

\newcommand*{\etal}{\emph{et al.}\@\xspace}

\makeatletter
\newcommand*{\etc}{%
    \@ifnextchar{.}%
        {etc}%
        {etc.\@\xspace}%
}

\definecolor{notetext}{rgb}{0.7,0,0}

\definecolor{ubpubColor}{rgb}{0.43, 0.5, 0.5}
\definecolor{backboneColor}{rgb}{0.423, 0.325, 0.365}
\definecolor{fpnColor}{rgb}{0.255, 0.498, 0.416}

\newcommand{\PAR}[1]{\vskip4pt \noindent {\bf #1~}}

\newcolumntype{P}[1]{>{\centering\arraybackslash}p{#1}}

\newcommand{\abbrev}{4D-StOP}

\begin{document}
\pagestyle{headings}
\mainmatter
\def\ECCVSubNumber{285}  

\title{\abbrev: Panoptic Segmentation of 4D LiDAR using Spatio-temporal Object Proposal Generation and Aggregation}

\titlerunning{\abbrev}
%
\author{Lars Kreuzberg\inst{1} \and
Idil Esen Zulfikar\inst{1} \and
Sabarinath Mahadevan\inst{1} \and
\\Francis Engelmann\inst{2} \and
Bastian Leibe\inst{1}
}
\authorrunning{L. Kreuzberg et al.}
%
\institute{$^1$RWTH Aachen University, Germany \quad $^2$ETH Zurich, AI Center, Switzerland
\email{lars.kreuzberg@rwth-aachen.de \\ \{zulfikar,mahadevan,leibe\}@vision.rwth-aachen.de \\
 francis.engelmann@ai.ethz.ch}}
\maketitle

\begin{abstract}

In this work, we present a new paradigm, called \abbrev{},
to tackle the task of 4D Panoptic LiDAR Segmentation.
{\abbrev{}} first generates spatio-temporal proposals using voting-based center predictions,
where each point in the 4D volume votes for a corresponding center.
These tracklet proposals are further aggregated using learned geometric features.
The tracklet aggregation method effectively generates a video-level 4D scene representation over the entire space-time volume.
This is in contrast to existing end-to-end trainable state-of-the-art approaches which use spatio-temporal embeddings
that are represented by Gaussian probability distributions.
Our voting-based tracklet generation method followed by geometric feature-based aggregation generates significantly improved panoptic LiDAR segmentation quality when compared to modeling the entire 4D volume using Gaussian probability distributions.
\abbrev{} achieves a new state-of-the-art when applied to the SemanticKITTI test dataset with a score of 63.9 LSTQ,
which is a large (+7\%) improvement compared to current best-performing end-to-end trainable methods.
The code and pre-trained models are available at: \texttt{\url{https://github.com/LarsKreuzberg/4D-StOP}}. 

\end{abstract}


\section{Introduction}


In recent years, we have made impressive progress in the field of 3D perception,
which is mainly due to the remarkable developments in the field of deep learning.
With the availability of RGB-D sensors, we have made progress on tasks like ~\cite{Qi2017PointNetDL,3dsemseg_ICCVW17,Landrieu2018LargeScalePC,3dsemseg_ECCVW18},
3D instance segmentation~\cite{Hou20193DSIS3S,Yang2019LearningOB,Lahoud20193DIS,Elich20193DBEVISBI,Engelmann20203DMPAMA, Chibane22ECCV} and
3D object detection~\cite{qi2019deep,Zhou2018VoxelNetEL,Misra2021AnET} which work on indoor scenes. 
Similarly, the accessibility to modern LiDAR sensors has made it possible to work with outdoor 3D scenes, where again recent works have targeted the tasks of 3D object detection~\cite{Yan2018SECONDSE,Shi2019PointRCNN3O,Yang2020ssd,Shi2020PVRCNNPF}, semantic segmentation~\cite{thomas2019kpconv,milioto2019iros,Zhang2020PolarNetAI,Zhu2021CylindricalAA}, panoptic segmentation~\cite{behley2021benchmark,Milioto2020LiDARPS,Zhou2021PanopticPolarNetPL,Hong2021LiDARbasedPS} and multi-object tracking~\cite{Weng2019ABF,Chiu2020Probabilistic3M,yin2021center,Bai2022TransFusionRL}. Perceiving outdoor 3D environments from LiDAR data is particularly relevant for robotics and autonomous driving applications, and hence has gained significant traction in the recent past.


\begin{figure}[ht!]
    \centering
    \includegraphics[width=1.0\linewidth]{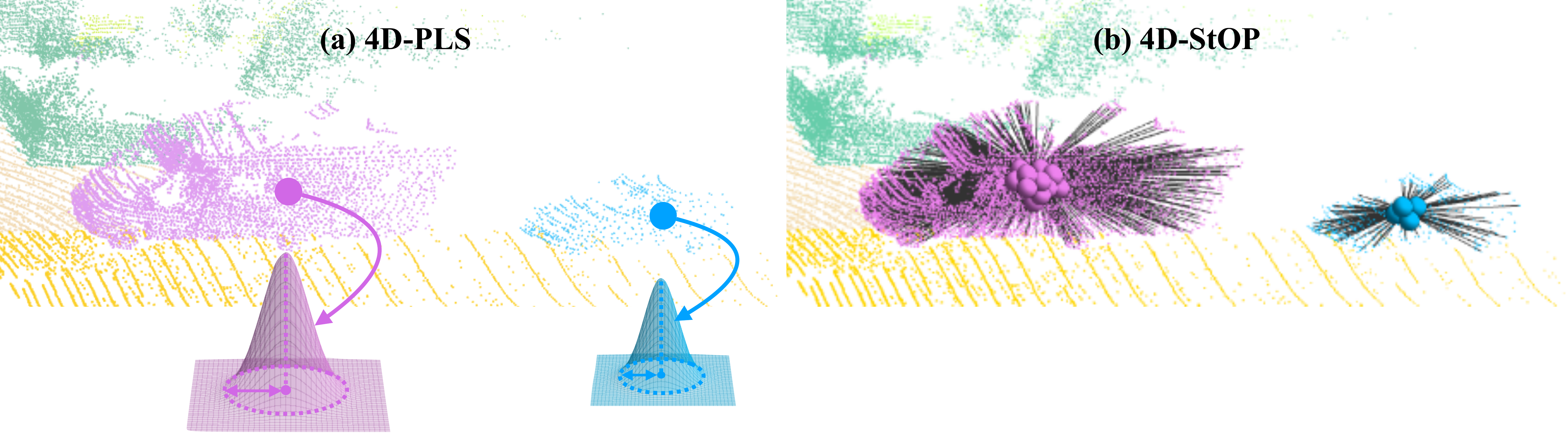}
    \caption{\textbf{Comparing 4D-PLS to {\abbrev{}}.}
    \textbf{(a)} The current end-to-end trainable state-of-the-art method models each tracklet as a Gaussian probability distribution in the 4D space-time volume.
    \textbf{(b)} Our work, 4D Spatial-temporal Object Proposal Generation and Aggregation (\abbrev), represents each tracklet with multiple proposals generated using voting-based center predictions in the 4D space-time volume, which are then aggregated to obtain the final tracklets.}
    \label{fig:probability-vs-voting}
\end{figure}

In the wake of these advancements, Ayg\"{u}n~\etal~\cite{aygun2021cvpr} proposed the 4D Panoptic LiDAR Segmentation task, where the goal is to perform panoptic segmentation of LiDAR scans across space-time volumes of point cloud data. Hence, each LiDAR point is assigned a semantic class label and a temporally consistent instance ID in a given sequence of 3D LiDAR scans, this task is more challenging than its predecessor 3D Panoptic LiDAR Segmentation~\cite{behley2021benchmark} since temporally consistent instance IDs and semantic class labels have to be preserved over the time continuum.
It is also critical for domains such as autonomous driving,
where autonomous agents have to be able to continuously interact with 4D environments and be robust to rapid changes in the space-time volume. 
%
%
%

For this task, the current research can be categorized into two different groups. In the first group, the existing methods follow a tracking-by-detection paradigm where an off-the-shelf 3D panoptic segmentation network~\cite{milioto2019iros,thomas2019kpconv,Zhu2021CylindricalAA} is first employed to obtain the instances in each scan along with the corresponding semantic class labels. The instances from different scans are then further assigned a consistent ID with the help of various data association methods~\cite{Weng2019ABF,Mittal_2020_CVPR,Lang2019PointPillarsFE,marcuzzi2022ral} in a second step.
Since an off-the-shelf network is used, these methods are not end-to-end trainable, and often utilize multiple different networks. 

The second group of methods~\cite{hurtado2020mopt,aygun2021cvpr} uses a single end-to-end trainable network to directly generate the semantic class label for each point and a temporally consistent instance ID for the corresponding objects.
The current top performing method~\cite{aygun2021cvpr} takes as input a 4D space-time volume which is generated by combining multiple LiDAR scans. A semantic segmentation network assigns  a semantic class to each of the input points, while the object instances are modeled as Gaussian probability distributions over the foreground points.
In addition to the Gaussian parameters, the network also predicts object centers and the tracklets are generated by clustering the points around these centers in the features space, by evaluating a Gaussian probability function as seen in Fig.~\ref{fig:probability-vs-voting}(a).
While such clustering techniques generally work well on small 4D volumes,
the method generates only one cluster center and therefore generates only one proposal for each object which limits the representation quality.


In this work, instead of representing tracklets as Gaussian distribution, we follow a new paradigm for forming tracklets in 4D space-time volume. Inspired by~\cite{Engelmann20203DMPAMA}, we represent tracklets as spatio-temporal object proposals, then aggregate these proposals to obtain the final tracklets as shown in Fig.~\ref{fig:probability-vs-voting}(b). 
Object proposals have been used successfully for object detection and instance segmentation in both 2D~\cite{ren15NIPS,he17iccv} and 3D~\cite{qi2019deep,Engelmann20203DMPAMA} domains. The success of these methods relies on producing a large number of proposals, thereby producing reliable outputs. 
In our paper, we investigate whether the success of using object proposals can also be observed in a 4D space-time volume, and empirically prove that they in fact do. For this, each feature point in our method votes for the closest center point to first generate object proposals, which are then aggregated using high-level learned geometric features in the 4D volume. For the first step, we utilize Hough voting~\cite{qi2019deep} in the 4D volume and show that it works well in this large and dynamic space. An alternative strategy to generate multiple proposals would be the adaptation of the Gaussian probability-based clustering, successfully used in both 2D~\cite{Athar_Mahadevan20ECCV,neven2019instance} and 3D~\cite{aygun2021cvpr}. However, we show that the performance of such an adaptation is inferior to Hough voting used in this work (see Sec.~\ref{subsec:gaussion-vs-voting}).  

Our method, 4D Spatio-temporal Object Proposal Generation and Aggregation (4D-StOP), generates better tracklets associations over an input 4D space-time volume, and correspondingly outperforms the state-of-the-art methods on the 4D Panoptic LiDAR Segmentation task.
In summary, we make \textbf{the following contributions}: (i) we tackle the task of 4D Panoptic LiDAR Segmentation in an end-to-end trainable fashion using a new strategy that first generates tracklet proposals and then further aggregates them; (ii) our method uses a center based voting technique to first generate tracklet proposals, which are then aggregated using our novel learned geometric features; (iii) we achieve the state-of-the-art results among all methods on SemanticKITTI test set, where we outperform the best performing end-to-end trainable method by a large margin. 
  
 
  
  

\section{Related Work}

\PAR{Point Cloud Segmentation.}
With the availability of multiple large-scale outdoor datasets~\cite{behley2019iccv,fong2021panoptic}, deep learning models are leading the field of LiDAR semantic segmentation ~\cite{thomas2019kpconv,Tang2020SearchingE3,Zhu2021CylindricalAA,wu2018SqueezeSegCN,milioto2019iros,cortinhal2020salsanext,Zhang2020PolarNetAI}.
These end-to-end trainable methods can be categorized into two groups: In the first group, the methods\cite{wu2018SqueezeSegCN,milioto2019iros,cortinhal2020salsanext,Zhang2020PolarNetAI} convert 3D point clouds to 2D grids and make use of 2D  convolution networks.
PolarNet\cite{Zhang2020PolarNetAI} projects 3D points to a bird's-eye-view and works on a polar coordinate system.  RangeNet++\cite{milioto2019iros} relies on a spherical projection mechanism and exploits range images.
The second group of methods\cite{thomas2019kpconv, Tang2020SearchingE3,Zhu2021CylindricalAA} directly works on 3D data to preserve the 3D geometry by using 3D convolution networks.
Cylinder3D~\cite{Zhu2021CylindricalAA} uses cylindrical partitioned coordinate systems and asymmetrical 3D convolution networks.
Similar to \cite{aygun2021cvpr}, we rely on KPConv~\cite{thomas2019kpconv} as feature backbone which directly applies deformable convolutions on point clouds.

\PAR{Multi-Object Tracking.} In 2D multi-object tracking (MOT), earlier methods~\cite{Milan2015JointTA,LealTaix2014LearningAI,Bergmann2019TrackingWB,Braso2020CVPR} follow the tracking-by-detection paradigm where off-the-shelf 2D object detection networks~\cite{ren15NIPS,he17iccv} are utilized to obtain objects in the frames, then the objects across frames are associated with different data association methods.
Later methods\cite{zhou2020tracking,Pang2021QuasiDenseSL,Wang2021JointGraph,meinhardt2021trackformer} shifted to joint detection and association in a single framework.
Recently, 3D LiDAR MOT has become popular in the vision community, due to the impressive developments in 3D object detectors~\cite{ren15NIPS,Lang2019PointPillarsFE} and newly emerging datasets~\cite{Caesar2020nuScenesAM,Sun2020ScalabilityIP}.
The methods tackle 3D MOT either with tracking-by-detection paradigm~\cite{Weng2020_AB3DMOT_eccvw,Kim21ICRA} or with joint-detection-and-tracking paradigm~\cite{yin2021center,Bai2022TransFusionRL}, similar to the methods in 2D MOT.
All of these methods work on single scans, \ie, on the spatial domain, to track objects over time.
In contrast, our method works on a unified space and time domain, \ie, 4D volume formed by combining multi-scans, to localize and associate objects.


\PAR{Panoptic LiDAR Segmentation and Tracking.} 4D Panoptic LiDAR Segmentation~\cite{hurtado2020mopt, aygun2021cvpr, marcuzzi2022ral} is an emerging research topic that unifies semantic segmentation, instance segmentation and tracking of LiDAR point clouds into a single framework. The initial work proposes a unified architecture consisting of three heads that tackle semantic segmentation, instance segmentation and tracking tasks individually. The recent work~\cite{marcuzzi2022ral} obtains semantic labels and 3D object detections from an off-the-self panoptic segmentation network and adapts contrastive learning~\cite{oord2018representation,caron2020unsupervised,ChenK0H20} into tracking to associate the 3D detections. The state-of-the-art method~\cite{aygun2021cvpr} in the end-to-end trainable frameworks is a bottom-up approach that jointly assigns semantic labels and associates instances across time. It works on a space-time volume where tracklets are generated by evaluating Gaussian distributions.  
We direct future research towards end-to-end trainable models, our work aims to improve tracklet representation in space-time volume, and propose to tackle tracklets representation as multi-proposal generation and aggregation in 4D volume.

\begin{figure}[htp]
    \centering
    \includegraphics[width=1.0\linewidth]{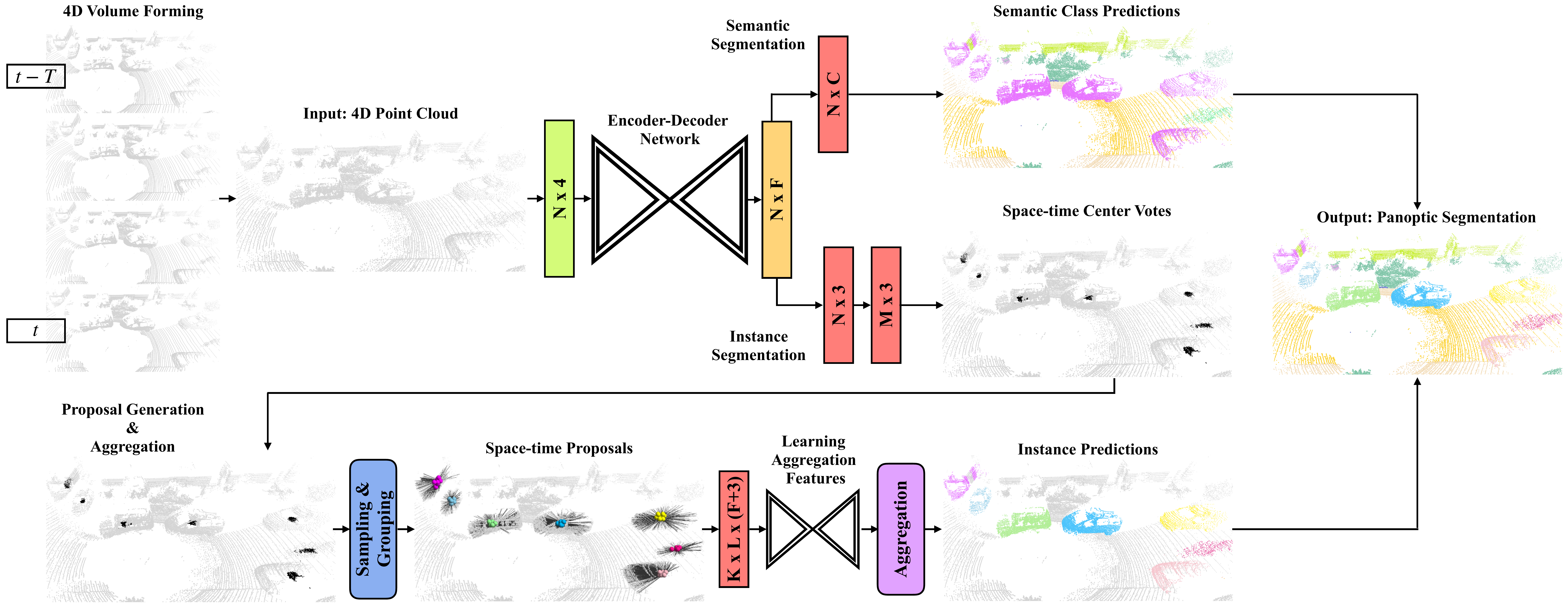}
    \caption{\textbf{\abbrev{} Overview.}
    Given a 4D point cloud, our method splits the task of 4D Panoptic LiDAR Segmentation into two subtasks semantic segmentation and instance segmentation. The semantic segmentation branch performs point-wise classification in 4D volume. The instance segmentation branch applies object-centric voting in 4D space-time to generate proposals followed by proposal aggregation based on learned features to obtain the final instances.
    $L$ is the number of points associated to a proposal.}
    \label{fig:4d-stop}
\end{figure}

\section{Method}
The overall architecture of our method is depicted in Fig.~\ref{fig:4d-stop}. We pose the task of 4D Panoptic LiDAR Segmentation by splitting it into the two subtasks, semantic segmentation as well as instance segmentation, and solve them in parallel. The semantic segmentation branch performs point-wise classification for each point in the 4D volume. For the instance segmentation task, we apply an object-centric voting approach in the 4D space-time volume. Based on the votes, multiple spatio-temporal proposals are generated. The features of these spatio-temporal proposals are learned to aggregate the proposals and generate the final tracklets. In the end, the results of the semantic segmentation and the instance segmentation branches are combined. To ensure that all points in an instance have the same semantic class label, we exploit majority voting where the most frequently occurring semantic class label among the instance points determines the semantic labels for all points within the instance.



\subsection{4D Volume Formation}
To form overlapping 4D point clouds, multiple consecutive scans are stacked together. For a scan at time-step $t$ and a temporal window size $T$, the point clouds within the temporal window $\{\max(0,t-T), ..., t\}$ are merged. However, stacking all the points from the scans in the temporal window $T$ is not possible due to memory constraints. To overcome this problem, we follow the \textit{importance sampling} strategy from~\cite{aygun2021cvpr}. Here, 10\% of the points with a probability proportional to the per-point objectness score are sampled from the previous scans at time-steps $t-T, ..., t-1$ while processing the scan at time-step $t$, which enables to focus on points from \textit{thing} classes while preserving the points from \textit{stuff} classes.

\subsection{4D-StOP}
\label{sec:4D-StOP}
\PAR{Semantic Segmentation and Voting for 4D Volume.}
Given a 4D point cloud with $N$ points as input, we first learn strong point features with a backbone that encodes the semantics and the geometry of the underlying multi-scan scene. We use an encoder-decoder architecture as our backbone~\cite{thomas2019kpconv} to generate per-point features $f_{i} \in \mathbb{R}^F$, where $i \in \{1,...,N\}$. On top of the backbone, we add a semantic head module and a voting module; and perform these tasks jointly. The semantic head predicts a semantic class label for each point and encodes the semantic context into the point features. To learn the semantic labels, a standard cross-entropy classification loss $L_{sem}$ is used. After the semantic head predicts a semantic class label for each point in the 4D volume, we distinguish between the points assigned to foreground and background classes and run the voting module only on the foreground points.  

In the voting module, we adapt the deep Hough voting used in~\cite{qi2019deep} to work on 4D space-time volumes. Every foreground point in the 4D volume votes for the corresponding space-time object center. In particular, this is realized by learning per-point relative offsets $\Delta x_i \in \mathbb{R}^3$ that point from the point positions $x_i \in \mathbb{R}^3$ to the corresponding ground truth object centers $c_i^* \in \mathbb{R}^3$ in the 4D space-time volume. The loss function for the voting module can be formulated as:
\begin{align}
    L_{vot} = \frac{1}{M}||x_i + \Delta x_i - c_i^*||_H \cdot \boldsymbol{1}(x_i), \label{eq: vot_loss}
\end{align}
where $|| \cdot ||_H$ is the Huber-Loss, $\boldsymbol{1} (\cdot)$ is a binary function indicating whether a point belongs to foreground, and $M$ is the number of points in foreground. In a multi-scan scenario, the ground truth bounding-box centers of the 3D objects, \ie,  the ground truth bounding-box centers on a single 3D scan, cannot be simply regressed in 4D volume. Since the objects move constantly in consecutive scans, it would lead to multiple regressed centers that would eventually yield insufficient results. To avoid that, we recompute the ground truth bounding-box centers for the objects in the 4D space-time volumes such that the ground-truth centers $c_i^*$ in Eq.~\ref{eq: vot_loss} are the ground-truth bounding-box centers of the objects in the merged 4D volume.

\PAR{Proposal Generation.} 
The proposals are generated by uniformly sampling and grouping the points based on the spatial-temporal proximity of the votes within a 4D volume. For each foreground point, there exists a vote to the corresponding space-time object centers. From these center votes, $K$ samples are chosen using the Farthest Point Sampling (FPS) selection technique used in~\cite{qi2019deep} to serve as proposal centers $y_j = x_j + \Delta x_j,  \forall j \in \{1,...,K\}$. As suggested in~\cite{Engelmann20203DMPAMA}, we also tested Random Sampling technique, which yields results on the same level as FPS. Although both sampling methods give similar results, we continue with the FPS technique as the probability of missing an object is much lower. To complete the proposal generation, all points are assigned to proposal $j$ if a point votes for a center within a radius $r$~($=0.6 m$) of the sampled proposal position $y_j$. In the end, each proposal contains the proposal positions $y_j$ near the space-time object centers and a set of points $s_{i}$ associated with each proposal $j$.

\PAR{Proposal Aggregation.}
\label{sec:aggfet}
A simple way to aggregate multiple proposals is to use the proposal center positions, \ie use relatively simple features. However, proposal aggregation can be enhanced by learning more advanced geometric features. We have $K$ object proposal in 4D space-time volume consisting of the proposal positions $y_j$ and a set of points $s_i$ associated with each proposal $j$. Also, each of the points associated with a proposal has a per-point feature $f_{i} \in \mathbb{R}^F$.
We get the proposal features $g_i \in \mathbb{R}^D$ which describes the local geometry of their associated objects by using a PointNet-like module\cite{Qi2017PointNetDL} applied to the per-point features of the associated points. In particular, the per-point features in combination with the per-point positions that belong to a proposal are processed by a shared MLP with output sizes $(128, 128, 128)$, then channel-wise max-pooling is applied to predict a single feature vector $g_i \in \mathbb{R}^D$. Each proposal is processed independently. The proposal features are further processed by another MLP with output sizes $(128, 128, E)$ to learn the $E$ dimensional geometric aggregation features.
Overall, we use three geometric features, where two of them, a refined proposal center point prediction $y_i + \Delta y_i$ and an object radius estimation $r_i$, are adapted from~\cite{Engelmann20203DMPAMA}. The third novel geometric feature that we use for aggregation is the predicted bounding box size $bb_i = (l_i, h_i, w_i) \in \mathbb{R}^3$. This additional geometric feature is helpful in better aggregating the proposal as shown in our experiments (see Sec.~\ref{sec:analysis}). The aggregation-loss is defined as: 
\begin{align}
    L_{agg} = ||y_i + \Delta y_i - c_i^*||_H + ||r_i - r_i^*||_H + ||bb_i - bb_i^*||_H
\end{align}
where $c_i^*$ is the center, $r_i^*$ is the radius of the closest ground truth bounding sphere and $bb_i^*$ are the length, height and width of closest ground truth  bounding-box.  
%
%
Finally, we aggregate the proposals by applying DBScan clustering to the learned geometric features. The proposals whose geometric features end up in the same cluster are combined. The union of the points in the aggregated proposals builds the final space-time object predictions, that is the final tracklets. 

\subsection{Tracking}
\label{subsec:tracking}

Within an input 4D volume, temporal association is resolved implicitly as object instances are grouped jointly in 4D space-time. Due to memory constraints, there is a limitation to the size of the 4D volume that can be processed at once. As a result, we split the entire 4D point cloud into multiple overlapping volumes of smaller sizes. Following the strategy used in~\cite{aygun2021cvpr}, final tracklets across the entire space-time volume are obtained by associating the individual fragmented tracklets based on the best IoU overlap scores. For IoU scores below a particular threshold, we discard the corresponding associations and start a new tracklet.


\subsection{Training and Implementation Details}
\label{sec:training_details}
The model is trained end-to-end using the multi-task loss $L = \alpha \cdot L_{sem} + \beta \cdot L_{vot} + \gamma \cdot L_{agg}$
 for 550K iterations in total. For the first 400K iterations, we set $\alpha = 1$, $\beta = 1$ and $\gamma = 0$ in order to predict the semantic classes and to obtain the spatio-temporal proposals in a sufficient quality. In the remaining 150K iterations, we set $\alpha = 0$, $\beta = 0$ and $\gamma = 1$ and freeze the semantic segmentation and the voting modules, then learn high-quality geometric features for aggregating the spatio-temporal proposals. Our model is implemented in PyTorch. The training is performed on a single NVIDIA A40 (48GB) GPU and the inference on a single NVIDIA TitanX (12GB) GPU. We apply random rotation, translation and scaling of the scene as data augmentations. Moreover, we randomly drop some points of the scene and randomly add noise.

\newcommand{\numberone}{\normalsize\ding{172}}%
\newcommand{\numbertwo}{\normalsize\ding{173}}%
\newcommand{\numberthree}{\normalsize\ding{174}}%
\newcommand{\numberfour}{\normalsize\ding{175}}%
\newcommand{\numberfive}{\normalsize\ding{176}}%

\section{Experiments}


\subsection{Comparing with State-of-the-Art Methods} 
\PAR{Dataset and Metric.}
We evaluate our method on the SemanticKITTI LiDAR dataset~\cite{behley2019iccv}. It consists of 22 sequences from KITTI odometry dataset~\cite{geiger2012cvpr}. There exist three splits: training, validation and test split where sequences 00 to 10 are used as training and validation split, and sequences 11 to 21 as test split. Each point is labeled with a semantic class and a temporally consistent instance ID. In total, there are 19 semantic classes among which 8 are \emph{things} classes and the remaining 11 are \emph{stuff} classes. The main evaluation metric is $LSTQ$, which is the geometric mean of the classification score $\mathrm{S}_{cls}$ and the association score $\mathrm{S}_{assoc}$, \ie \(\emph{LSTQ} = \sqrt{\emph{$S_{cls}$} \times \emph{$S_{assoc}$}}\)~\cite{aygun2021cvpr}.
The classification score $\mathrm{S}_{cls}$ measures the quality of semantic segmentation, while the association score $\mathrm{S}_{assoc}$ measures the quality of association in a 4D continuum. 
Unlike the previous evaluation metrics~\cite{Voigtlaender2019MOTSMT,Kim2020VideoPS,hurtado2020mopt}, \emph{LSTQ} performs semantic and instance association measurements independently to avoid penalization due to the entanglement of semantic and association scores.
Our strategy of generating proposals and then aggregating them in the 4D volume generates very good object associations while maintaining the semantic classification quality, which is reflected in the association ($\mathrm{S}_{assoc}$) and the classification scores ($\mathrm{S}_{cls}$) in the following experiments.


\begin{table}[t]
\renewcommand{\arraystretch}{1.2}
\center
\normalsize{
\tabcolsep=0.16cm
 \begin{tabular}{l l |c c c | c c} 
 \toprule
 & Method & \textit{LSTQ} & $\mathrm{S_{assoc}}$ & $\mathrm{S_{cls}}$ & $\mathrm{IoU^{St}}$ & $\mathrm{IoU^{Th}}$ \\
 \midrule
 \parbox[t]{1mm}{\multirow{5}{*}{\rotatebox[origin=c]{90}{\scriptsize{Not-End-to-End}}}} 
 & RangeNet++~\cite{milioto2019iros} + PP + MOT & 35.5 & 24.1 & 52.4 & 64.5 & 35.8 \\ 
 & KPConv~\cite{thomas2019kpconv} + PP + MOT & 38.0 & 25.9 & 55.9 & 66.9 & 47.7 \\
 & RangeNet++~\cite{milioto2019iros} + PP + SFP & 34.9 & 23.3 & 52.4 & 64.5 & 35.8 \\
 & KPConv~\cite{milioto2019iros} + PP + SFP & 38.5 & 26.6 & 55.9 & 66.9 & 47.7 \\
 & CIA~\cite{marcuzzi2022ral} & 63.1 &  65.7 &  \textbf{60.6} & 66.9 & 52.0 \\
 \midrule
 \parbox[t]{1mm}{\multirow{3}{*}{\rotatebox[origin=c]{90}{\scriptsize{End-to-End}}}} 
 & 4D-PLS (4 scan)~\cite{aygun2021cvpr}& 56.9 & 56.4 & 57.4 & 66.9 & 51.6 \\ 
 & \abbrev{}(Ours)(2 scan) & 62.9 &  67.3 & 58.8 & \textbf{68.3} & 53.3 \\
 & \abbrev{}(Ours)(4 scan) & \textbf{63.9} & \textbf{69.5} & 58.8 & 67.7 & \textbf{53.8} \\
 \bottomrule
 \end{tabular}

\caption{\textbf{Scores on SemanticKITTI 4D Panoptic Segmentation test set.} MOT - \emph{tracking-by-detection} by~\cite{Weng2019ABF}, SFP - \emph{tracking-by-detection} via scene flow based propagation~\cite{Mittal_2020_CVPR}, PP - PointPillars~\cite{Lang2019PointPillarsFE}.}
\label{tab:semantickitti-test-set}
}
\end{table}
\begin{table}[t]
\renewcommand{\arraystretch}{1.2}
\center
\normalsize{
\tabcolsep=0.16cm
 \begin{tabular}{l l |c c c | c c} 
 \toprule
 & Method & LSTQ & $\mathrm{S_{assoc}}$ & $\mathrm{S_{cls}}$ & $\mathrm{IoU^{St}}$ & $\mathrm{IoU^{Th}}$ \\
 \midrule
 \parbox[t]{1mm}{\multirow{4}{*}{\rotatebox[origin=c]{90}{\scriptsize{Not-End-to-End}}}} 
 & RangeNet++~\cite{milioto2019iros} + PP + MOT & 43.8 & 36.3 & 52.8 & 60.5 & 42.2 \\ 
 & KPConv~\cite{thomas2019kpconv} + PP + MOT & 46.3 & 37.6 & 57.0 & 64.2 & 54.1 \\
 & RangeNet++~\cite{milioto2019iros} + PP + SFP & 43.4 & 35.7 & 52.8 & 60.5 & 42.2 \\
 & KPConv~\cite{milioto2019iros} + PP + SFP & 46.0 & 37.1 & 57.0 & 64.2 & 54.1 \\
 \midrule
 \midrule
 \parbox[t]{1mm}{\multirow{5}{*}{\rotatebox[origin=c]{90}{\scriptsize{End-to-End}}}} 
 & MOPT~\cite{hurtado2020mopt} &  24.8 & 11.7 & 52.4 & 62.4 & 45.3 \\
 & 4D-PLS~\cite{aygun2021cvpr} (2 scan) & 59.9 & 58.8 & 61.0 & 65.0 &  63.1 \\ 
  & 4D-PLS~\cite{aygun2021cvpr} (4 scan) & 62.7 & 65.1 & 60.5 & \textbf{65.4} & 61.3 \\ 
  & \abbrev{}(Ours)(2 scan) & 66.4 & 71.8 & \textbf{61.4} & 64.9 & \textbf{64.1} \\
 & \abbrev{}(Ours)(4 scan) & \textbf{67.0}
 & \textbf{74.4} & 60.3 & 65.3 & 60.9 \\
 \bottomrule
 \end{tabular}

\caption{\textbf{Scores on SemanticKITTI 4D Panoptic Segmentation validation set.}
MOT - \emph{tracking-by-detection} by~\cite{Weng2019ABF}, SFP - \emph{tracking-by-detection} via scene flow based propagation~\cite{Mittal_2020_CVPR}, PP - PointPillars~\cite{Lang2019PointPillarsFE}.}
\label{tab:semantickitti-validation-set}
}
\end{table}

\PAR{Results.}
In Tab.~\ref{tab:semantickitti-test-set}, we present the results on the SemanticKITTI 4D Panoptic Segmentation test set. Among the end-to-end trainable methods, \abbrev{} outperforms the previous state-of-the-art method 4D-PLS~\cite{aygun2021cvpr} by a large margin, obtaining a \textbf{7\%} boost in \emph{LSTQ} score, when utilizing the same number of scans (4 scans). Especially, we achieve \textbf{13.1\%} improvement in $\mathrm{S}_{assoc}$, which shows that \abbrev{} produces better segmented and associated tracklets in a 4D space-time volume compared to 4D-PLS.
While we achieve state-of-the-art results when using 4 scans, \abbrev{} also performs well with only 2 scans. The performance drop in the \emph{LSTQ} score is merely \textbf{1\%}.

Comparing \abbrev{} with the non-end-to-end trainable methods, we observe that \abbrev{} also outperforms the current not-end-to-end trainable state-art-the-art method CIA~\cite{marcuzzi2022ral} (+\textbf{0.8 \textit{LSTQ}}). We would like to highlight that CIA leverages an off-the-shelf 3D panoptic segmentation network with a stronger backbone than used in \abbrev{}. Thus, it has a better $\mathrm{S}_{cls}$ score than \abbrev{}, \ie the increase in $\mathrm{S}_{cls}$ score results from the semantic segmentation performance of the backbone network.
In contrast to CIA, we do not take advantage of such a stronger backbone but rather preserve the exact same configuration of our baseline 4D-PLS. Despite the worse classification score, the \emph{LSTQ} score for \abbrev{} is still better than for CIA. Reason for this is the better working object association in our end-to-end trainable approach. \abbrev{} is better in $S_{assoc}$ than CIA by a large margin with \textbf{3.8\%} boosting, where the increase is gained from the method basically.

In Tab.~\ref{tab:semantickitti-validation-set}, we report the results on the SemanticKITTI 4D Panoptic Segmentation validation set. In the validation set results, we observe a similar pattern as in the test set. \abbrev{} outperforms all the methods by a large margin in both not-end-to-end trainable and end-to-end trainable categories. It produces better scores compared to 4D-PLS by \textbf{+4.3 \textit{LSTQ}}  and \textbf{+9.3 $\mathrm{S}_{assoc}$} for 4 scans, and by \textbf{+6.5 \textit{LSTQ}} and \textbf{+13.0 $\mathrm{S}_{assoc}$} for 2 scans.
Although there is a slight performance drop between the two and four scan set-ups, this drop is smaller compared to 4D-PLS (\textbf{0.6\% \textit{LSTQ}} drop for \abbrev{} compared to \textbf{2.8\% \textit{LSTQ}} drop for 4D-PLS).
This shows that \abbrev{} is more robust to the number of input scans.
Since processing two scans is faster than processing four scans,
\abbrev{} offers a better trade-off between accuracy and speed than 4D-PLS.

\subsection{Analysis}
\label{sec:analysis}
We ablate all of our design choices on the validation split of the SemanticKITTI dataset, \ie only on sequence 08 of the dataset.
In this section, we provide detailed experimental analysis, and discuss the impact of each of the components in our method, including the geometric features. In addition, we also reformulate \abbrev{}, where the proposals are modeled as Gaussian distributions similar to our baseline 4D-PLS, and compare this formulation with our voting based approach.
In the experimental analysis, a 4D volume is formed by combining 2 scans unless it is not explicitly indicated it is generated from 4 scans. 


 
\newcommand{\analysisone}{\large\ding{172}}%
\newcommand{\analysistwo}{\large\ding{173}}%
\newcommand{\analysisthree}{\large\ding{174}}%
\newcommand{\analysisfour}{\large\ding{175}}%
\newcommand{\analysisfive}{\large\ding{175}}%
\begin{table}[t]
\renewcommand{\arraystretch}{1.2}
\center
\normalsize{
\tabcolsep=0.10cm
 \begin{tabular}{l l | c c c} 
 \toprule
 {} & {} & \textit{LSTQ} &  $\mathrm{S_{assoc}}$ & $\mathrm{S_{cls}}$ \\
 \midrule
 \analysisone & 4D-PLS~\cite{aygun2021cvpr} (2 scan) & 59.9 & 58.8 & 61.0 \\ 
 \analysistwo & Our Baseline & {63.7} & {66.7} & {60.8} \\
 \analysisthree & Aggregating w/ Proposal Positions & {65.3} & {70.0} & {60.8} \\ 
 \analysisfour & Aggregating w/ Proposal Positions + Majority Voting & {65.5} & {70.0} & {61.2} \\ 
 \analysisfive & Aggregating w/ Geometric Features + Majority Voting & {\textbf{66.4}} & {\textbf{71.8}} & {\textbf{61.4}} \\ 
 \bottomrule
 \end{tabular}

\caption{\textbf{Analysis of each component in \abbrev{}.}}
\label{tab:ablate-components}
}
\end{table}

\PAR{Effect of \abbrev{} components.}
\label{subsec:effect_components}
We evaluate the impact of each \abbrev{} component in Tab.~\ref{tab:ablate-components}. The main improvements come from the proposal generation and aggregation steps. 
To investigate the impact of the aggregation step, we create a baseline in experiment \numbertwo{} by applying the traditional non-maximum-suppression (NMS) on the spatio-temporal proposals.
Even this baseline outperforms 4D-PLS by a large margin with \textbf{3.8\%} boosting in the \textit{LSTQ} score.
In order to see the effect of the aggregation step, we first conduct a naive aggregation by combining proposals based on the proposal positions (experiment \numberthree) and observe another \textbf{1.6\%} increase in the \textit{LSTQ} score.
In experiment \numberfour, we ablate adding majority voting into our pipeline which only brings a slight increment (\textbf{+0.4 $\mathrm{S}_{cls}$}).
Lastly, in experiment \numberfive{}, we show the results for adding the learned high-level geometric features as the final component, which boosts the performance in the $\mathrm{S}_{assoc}$ and \textit{LSTQ} scores by \textbf{1.8\%} and \textbf{0.9\%} respectively.
Overall, our method outperforms the prior end-to-end trainable method 4D-PLS~\cite{aygun2021cvpr} significantly irrespective of whether we use our proposal aggregation strategy or a much simpler NMS.
This shows that, similar to the observations made in 2D and 3D spatial domains, proposal-based methods surpass other bottom-up approaches in the 4D space-time domain.


\PAR{\abbrev{} with Gaussian Distribution.}
An alternative way of generating proposals would be to use Gaussian probability distributions. To do this, we adapt 4D-PLS~\cite{aygun2021cvpr} to work with our proposal generation and aggregation paradigm. In all of the following experiments, 4 scan setup is used to be consistent with the best results in 4D-PLS. We generate proposals based on per-point objectness scores, which indicate the probability that a point is close to an object center. We select all the points in the 4D volume with an objectness score above a certain threshold as the proposal center points.
Then, we assign points to a proposal by evaluating each point under the Gaussian probability distribution function with the proposal center as cluster seed point. For each point, the Gaussian probability indicates whether the point belongs to the proposal.
All the points with a probability above a certain threshold are assigned to the proposal. We generate the proposals in parallel, and thus obtain overlapping Gaussian probabilities for each object, \ie the multiple proposals for each object. In contrast, in 4D-PLS each object is represented with one proposal. After the proposals are generated, they are aggregated using the cluster mean points.
In Tab.~\ref{tab:ablate-4dstop-gaussian}, we compare 4D-PLS and \abbrev{} with Gaussian distribution (experiments \numbertwo{} and \numberthree). \abbrev{} with Gaussian distribution yields worse results than 4D-PLS (\textbf{-1.5 \textit{LSTQ}}).
We realized that using all proposals with an objectness score above a threshold is prone to produce more FPs than 4D-PLS, so we decided to increase the probability threshold while assigning each point to the proposals. A slight boosting \textbf{+0.7 \textit{LSTQ}} is achieved in this experiment (the experiment \numberfour), but it is still largely under our work \abbrev{}. To eliminate more FPs, we consider another strategy, where only a subset of the proposals is selected for aggregation. After generating the proposals, we start selecting the proposal with the highest objectness score from the proposals candidate pool. The proposals within a certain radius around the center point of the selected proposal are removed from the candidate pool. We repeat it until no proposal is left in the candidate pool and only aggregate the selected proposals in the end. With this strategy, a minor boosting by \textbf{+0.4 \textit{LSTQ}} is observed compared to 4D-PLS in experiment \numberfive, but the results are still under \abbrev{}.

In conclusion, we show that the multi-proposal generation and aggregation strategy can be implemented with Gaussian probability-based clustering used in 4D-PLS, but this does not bring a significant improvement to the base method. More importantly, the performance of such an adaptation still falls below \abbrev{} by a large margin.
Thereby, with these experiments we show that Gaussian distributions can be used to represent tracklet proposals, although the performance improvement brought by them saturates soon, and are inferior to our center-based voting approach.

\label{subsec:gaussion-vs-voting}

\newcommand{\numone}{\large\ding{172}}%
\newcommand{\numtwo}{\large\ding{173}}%
\newcommand{\numthree}{\large\ding{174}}%
\newcommand{\numfour}{\large\ding{175}}%
\newcommand{\numfive}{\large\ding{176}}%
\begin{table}[t]
\renewcommand{\arraystretch}{1.2}
\center
\normalsize{
\tabcolsep=0.20cm
 \begin{tabular}{l l | c c | c c c} 
 \toprule
 {} & {} & {PT} & {R}& \textit{LSTQ} &  $\mathrm{S_{assoc}}$ & $\mathrm{S_{cls}}$ \\
 \midrule
 \numone & \abbrev{}(Ours) & {-} & {-} & \textbf{67.0} & \textbf{74.4} & 60.3 \\
 \numtwo & 4D-PLS & {0.5} & {-} & {62.7} & {65.1} & {\textbf{60.5}} \\
 \numthree & \abbrev{} w/ Gaussian Distribution & {0.5} & {0.0} & {61.2} & {62.1} & {60.3} \\ 
 \numfour & \abbrev{} w/ Gaussian Distribution & {0.7} & {0.0} & {63.4} & {66.7} & {60.3} \\ 
 \numfive & \abbrev{} w/ Gaussian Distribution & {0.7} & {0.6} & {63.1} & {66.0} & {60.3} \\ 
 \bottomrule
 \end{tabular}

\caption{\textbf{Analysis for studying \abbrev{} with Gaussian Distribution.} PT - \textit{Probability Threshold} and R - \textit{Radius}. All of the experiments are carried out on 4 scans and the objectness score threshold is set to 0.7 as in 4D-PLS.}
\label{tab:ablate-4dstop-gaussian}
}
\end{table}


\newcommand{\cmark}{\ding{51}}%
\newcommand{\xmark}{\ding{55}}%
\begin{table}[t]
\renewcommand{\arraystretch}{1.2}
\center
\normalsize{
\tabcolsep=0.16cm
 \begin{tabular}{c c c c c c c} 
 \toprule
 \multicolumn{4}{c}{\textbf{Geometric Features}} & \multirow{2}{*}{\textit{LSTQ}} &  \multirow{2}{*}{$\mathrm{S_{assoc}}$} & \multirow{2}{*}{$\mathrm{S_{cls}}$} \\
 \cmidrule{1-4}
 Center Pos. & Ref. Center Pos. & Radius & Bounding-Box & {} & {} & {} \\
 \midrule
 \cmark & {--} & {--} & {--} & 65.5 & 70.0 & 61.2 \\
 {--} & \cmark & {--} & {--} & 65.8 & 70.7 & 61.3 \\
 {--} & \cmark & \cmark & {--} & 65.9 & 70.9 & 61.3 \\
 {--} & \cmark & \cmark & \cmark & \textbf{66.4} & \textbf{71.8} & \textbf{61.4} \\
 \bottomrule
 \end{tabular}

\caption{\textbf{Analysis of the learned geometric features.}}
\label{tab:ablate-aggregaion-loss}
}
\end{table}

\PAR{Impact of the learned geometric features.}
Our \abbrev{} method makes use of three different geometric features: the refined proposal centers, the radius estimation and the predicted bounding-box size (Sec.~\ref{sec:aggfet}), to aggregate the space-time proposals. We evaluate the impact of each geometric feature on the results of our method in Tab.~\ref{tab:ablate-aggregaion-loss}. Grouping proposals based on the proposals center positions (first row), \ie grouping without utilizing the learned geometric features, serves as our baseline. First, we analyze the performance for two learned geometric features, the refined center positions and the radius estimation, which we adapt from~\cite{Engelmann20203DMPAMA}. When the refined center positions are used individually (second row) or are combined with the radius estimation (third row), a performance increase can be observed. However, we achieve a significant performance boost by introducing the bounding-box size as an additional geometric feature (fourth row). We obtain \textbf{+1.8} and \textbf{+0.9} increase on the $\mathrm{S}_{assoc}$ and \emph{LSTQ} scores, respectively. The improvement is doubled by the additional use of the predicted bounding-box sizes compared to only use the refined proposal center positions and the radius estimation as the geometric features. By that, we prove two aspects. Our idea, introducing new geometric features, brings a notable boosting. Moreover, the quality of the learned geometric features affects the performance.

\begin{figure}[ht!]
    \centering
    \includegraphics[width=1.0\linewidth]{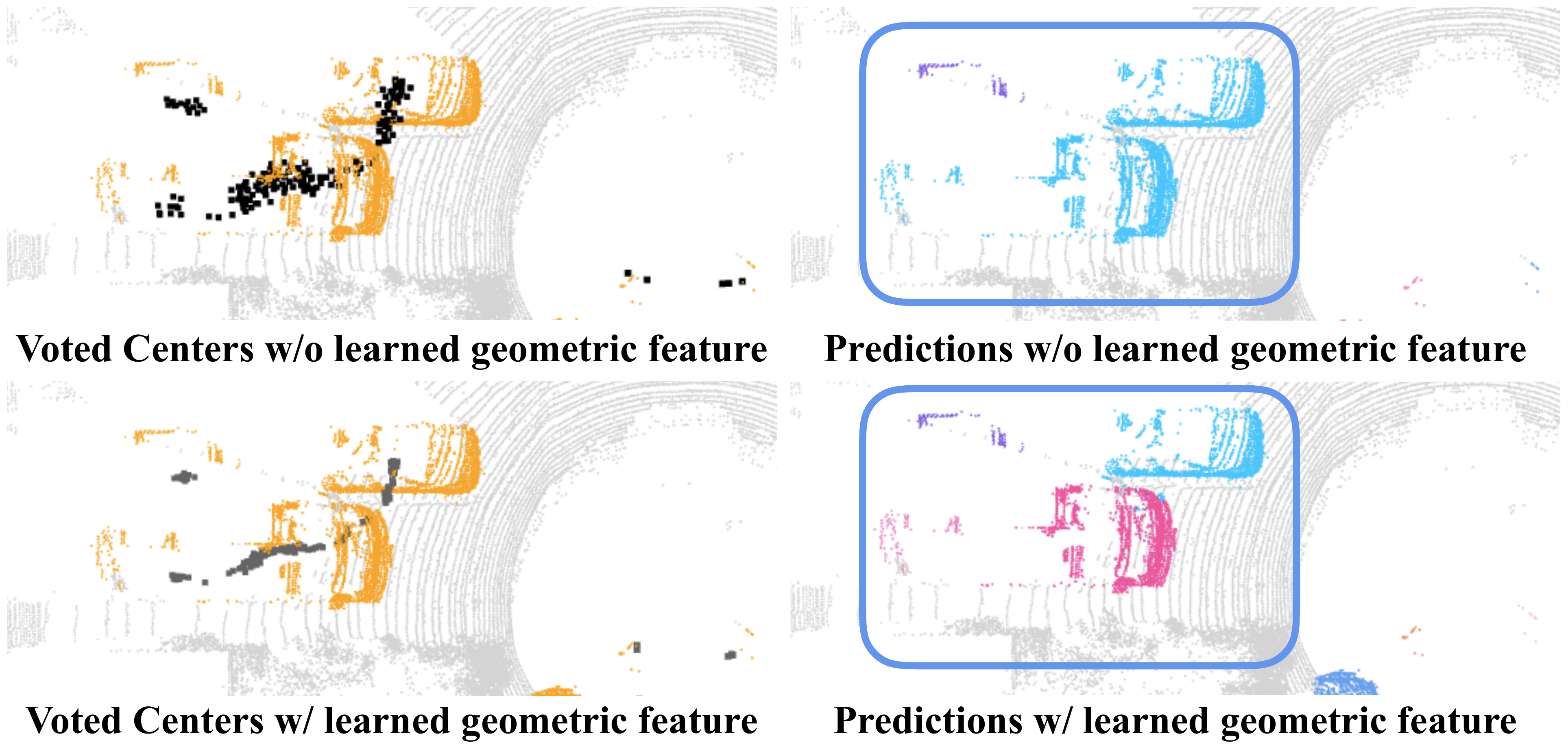}
    \caption{\textbf{Learned Geometric Features.}
    On an example scan, we show the impact of a learned geometric feature, the refined space-time proposal center point positions. In the first column, we display how the center positions change in 4D space-time volume with the learned geometric feature. More precise center positions can be observed using the learned geometric features. In the second column, the predictions are depicted with and without the learned geometric feature. Inside the blue boxes, it can be seen how a wrong prediction is corrected by using the learned geometric feature. In order to emphasize the results, the points with the \emph{stuff} class label are colored gray. The example is generated by combining two scans.}
    \label{fig:learned-geo-features}
\end{figure}

The impact of the learned geometric features is also visualized in an example scan in Fig.~\ref{fig:learned-geo-features}. The refined center positions are used as the geometric feature. In the first column, it is possible to observe how the proposal center positions get more precise by the use of the learned refinement. In the second row, the refined predicted proposal centers are closer to the ground truth centers compared to the predicted proposal centers without refinement in the first row. Also, we show how the predictions change with the learned geometric features in the second column. As seen in the blue boxes, we can correctly separate two objects predicted as one single object without the learned geometric features.

\subsection{Qualitative Results}
\begin{figure}[t]
    \centering
    \includegraphics[width=1.0\linewidth]{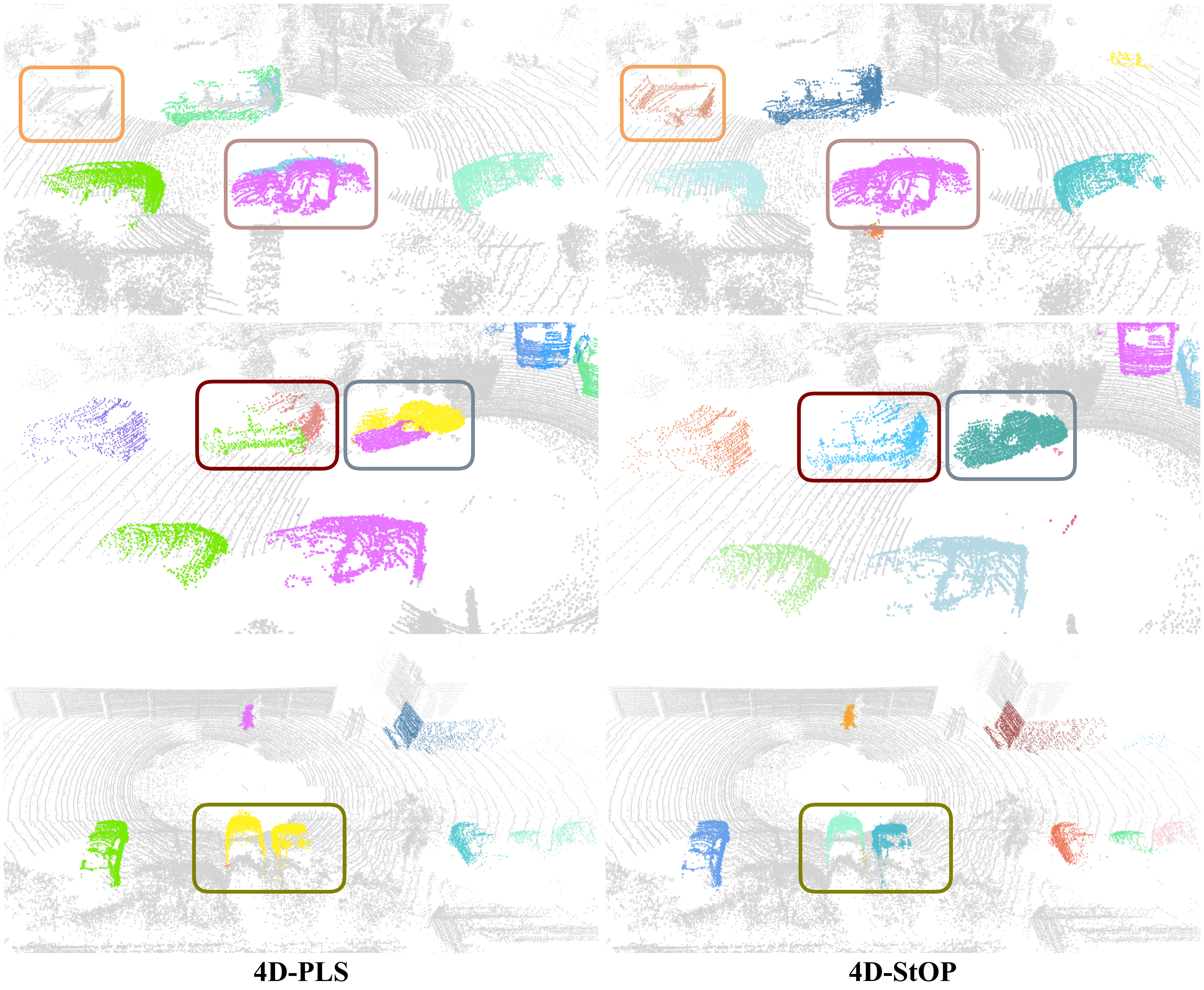}
    \caption{\textbf{Qualitative results on SemanticKITTI validation.}
    We show results for 4D-PLS {(left)} and \abbrev{} {(right)} of the same scans.
    Interesting cases are highlighted with boxes.
    Our method catches missed objects (first row),
    complete partially segmented objects (second row),
    and separate two incorrectly predicted objects as one single object (third row).
    For simplicity, points with \emph{stuff} class label are colored gray.
    All the scans are 4D volumes over 4 scans.}
    \label{fig:qualitative}
\end{figure}

In Fig.~\ref{fig:qualitative}, we present qualitative results for different example scans in the first and second column for 4D-PLS and our \abbrev{} method respectively.
We indicate each specific case within the boxes.
Compared to 4D-PLS, our method is able to catch and segment the missing object (left example in the first row), complete the partly segmented objects (right example in the first row and the examples in the second row), and correctly separate the two objects predicted as one single object (the middle example in the third row). 

\begin{figure}[ht!]
    \centering
    \includegraphics[width=1.0\linewidth]{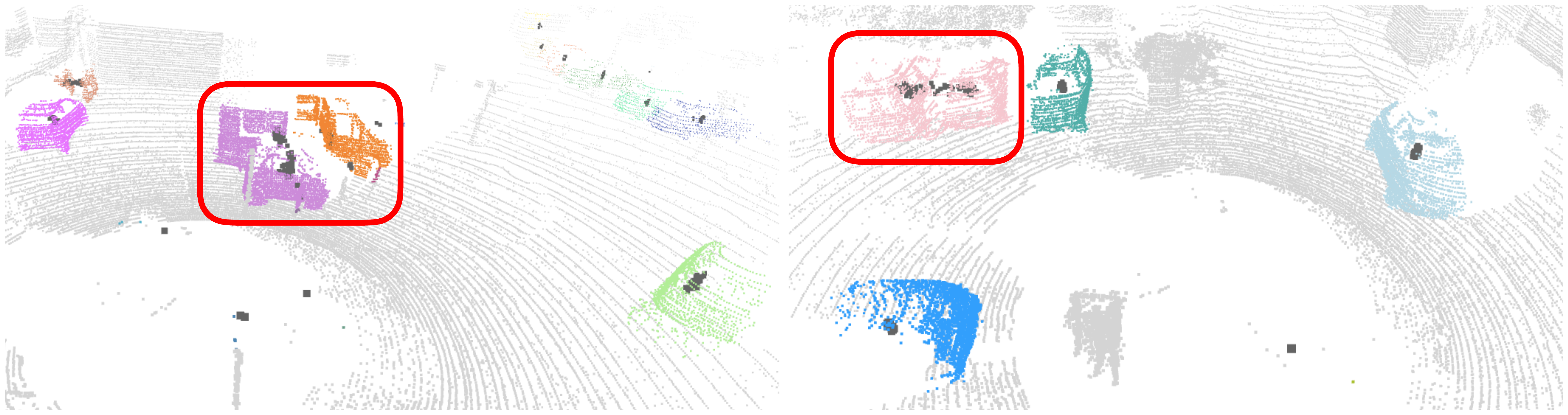}
    \caption{\textbf{Failure cases of \abbrev{} on the SemanticKITTI validation split.} 
    We show two scans where \abbrev{} fails.
    Here, \abbrev{} cannot separate two objects and mixes them or predicts them as one single object (see red boxes).
    The dark gray squares show the refined proposal centers. Again, the points with \emph{stuff} class label are colored gray and all the scans are 4D volumes over 4 scans.}
    \label{fig:failure}
\end{figure}

Fig.~\ref{fig:failure} shows two examples where \abbrev{} fails (red boxes).
We observe that \abbrev{} can fail when objects and thus their object center votes are very close to each other and objects have a similar size.
Then, the aggregation features of the corresponding proposals are quite similar and it is hard for \abbrev{} to distinguish which proposal belongs to which object. As a consequence, \abbrev{} mixes the objects (left) or predicts the objects as one instance (right).


\section{Conclusion}
In this work, we have proposed a new end-to-end trainable method \abbrev{} for 4D Panoptic LiDAR Segmentation task, which models tracklets in a space-time volume by generating multiple proposals, then aggregating them. Our model is based on center voting for generating tracklet proposals and learning high-level geometric features for aggregating tracklet proposals. We have introduced novel geometric features to enhance the proposal aggregation that improves performance notably. Additionally, we have proved that center voting is feasible in the space-time volume. We achieve state-of-the-art results on SemanticKITTI test-set. We hope that our work would inspire new representations of tracklets for future methods tackling 4D Panoptic LiDAR Segmentation task in a unified space-time volume.

\begin{small}{\PAR{Acknowledgments.}
We thank Sima Yagmur Zulfikar for her help and feedback on the figures, and Istv{\'a}n S{\'a}r{\'a}ndi for his helpful comments on our manuscript. This project was funded by ERC Consolidator Grant DeeVise (ERC-2017-COG-773161). The computing resources for several experiments were granted by RWTH
Aachen University under project 'supp0003'. Francis Engelmann is a post-doctoral research fellow at the ETH AI Center. This work is part of the first author's master thesis.}
\end{small}

\clearpage
%
%
\bibliographystyle{splncs04}
\bibliography{abbrev_short, 285}

\clearpage


\begin{center}
\Large\textbf{Supplementary Material} \\
\end{center}


\appendix

\section{Additional Ablation Studies}
\label{sec:additional_ablation}

In order to provide better insight, we present additional ablation studies on the validation split of the SemanticKITTI dataset. In this section, we compare the run-time performance of our method \abbrev{} with 4D-PLS. We also examine how the number of generated proposals and the radius size of generated proposals affect the results. Moreover, we discuss utilizing the random sampling technique over the Farthest Point Sampling (FPS) technique for selecting $K$ proposal centers. In all experimental analyses, we form a 4D volume by combining 2 scans unless it is not explicitly indicated it is generated from 4 scans.

\begin{figure}[ht!]
    \centering
    \includegraphics[width=1\linewidth]{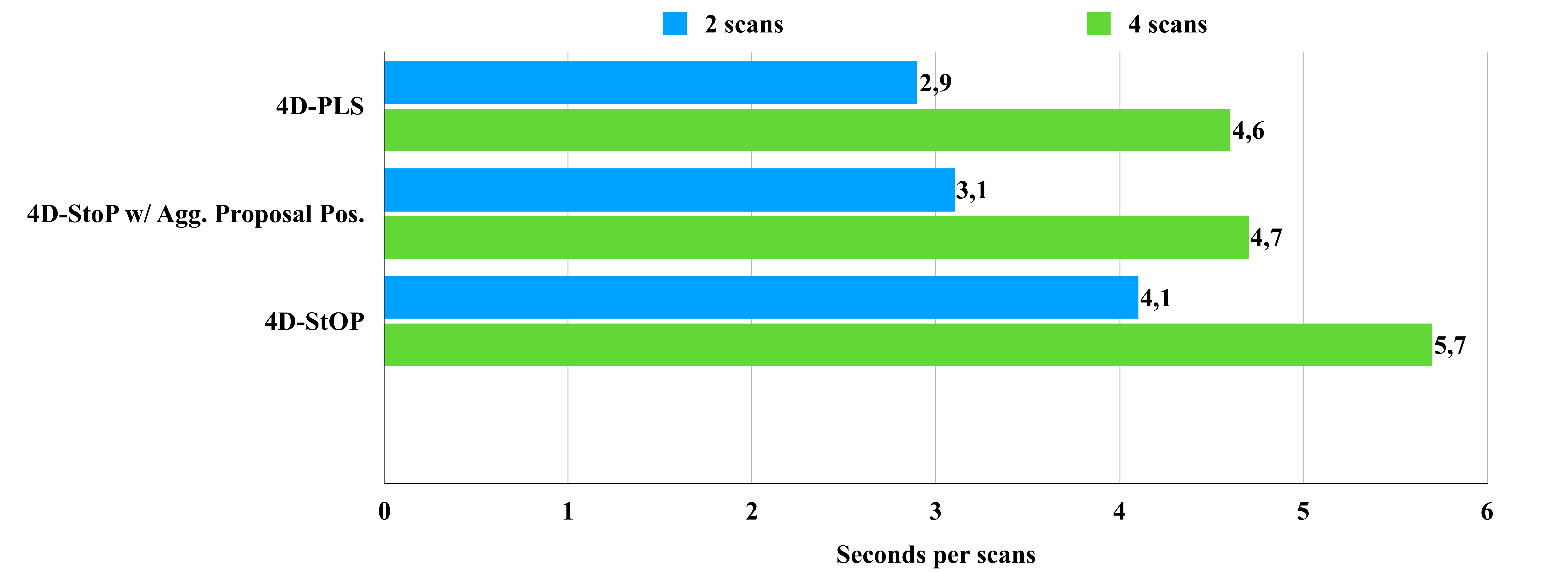}
    \caption{\textbf{Run-time performance comparison of 4D-PLS and \abbrev{}.} The results are depicted for 2 scan and 4 scan versions. Times are scans per second (sps).}
    \label{fig:runtime}
\end{figure}

\PAR{Run-time Analysis.}
We show run-time analysis for 4D-PLS and \abbrev{} in Fig.~\ref{fig:runtime}. We analyze the results for 2 scan and 4 scan setup separately. Times are recorded as scans per second (sps). We initially compare the run-time performance of 4D-PLS with a simpler version of \abbrev{} where the proposals are combined based on proposal positions, \ie without having the network module for learning aggregation features. For this case, we have an on-par run-time performance with 4D-PLS for both 2 scan and 4 scan setups, the difference is insignificantly small ($\sim$ 0.2 sps). We would like to remind this simpler version of \abbrev{} outperforms 4D-PLS in the \textit{LSTQ} score by \textbf{5.4\%} in 2 scan setup while having the on-par run-time (see Sec~\ref{subsec:effect_components}), \ie this simpler version offers a much better accuracy than 4D-PLS without losing any speed. If we add the network module for learning aggregation features, \ie when we have the complete version of our method, \abbrev{} is 1.2 and 1.1 sps slower than 4D-PLS for 2 scan and 4 scan setups, respectively. The run-time performance drops basically due to the network module for learning aggregation features. 
Note that, all of these run-time analyses are performed on a single Nvidia TitanX (12GB) GPU.

\begin{center}
\begin{table}[t]
\renewcommand{\arraystretch}{1.2}
\center
\normalsize{
\tabcolsep=0.10cm
 \begin{tabular}{c | c c c} 
 \toprule
 {\# Proposals} &\textit{LSTQ} &  $\mathrm{S_{assoc}}$ & $\mathrm{S_{cls}}$ \\
 \midrule
100  & {65.6} & {70.1} & {61.4} \\ 
200  & {66.4} & {71.7} & {61.4} \\ 
300  & {66.3} & {71.5} & {61.4} \\ 
400  & {66.3} & {71.7} & {61.4} \\ 
\textbf{500}  & \textbf{66.4} & \textbf{71.8} & \textbf{61.4} \\ 
600 & {66.3} & {71.7} & {61.4} \\ 
 \bottomrule
 \end{tabular}
\caption{\textbf{Ablation study on the impact of the number of proposals in \abbrev{}.}  Proposals are aggregated based on the learned geometric features and the radius size of proposals is set to 0.6$m$.}
\label{tab:ablate-num-prop}
}
\end{table}
\end{center}

\PAR{Effect of the number of proposals.}
In Tab.~\ref{tab:ablate-num-prop}, we evaluate the impact of the number of generated proposals. As seen, our method already works well with only 200 proposals. However, the number of proposals should not be set too low. With 100 proposals, we lose $1.6\%$ in $\mathrm{S}_{assoc}$ and $0.8\%$ in $LSTQ$ compared to the results with 200 proposals. The fewer proposals, the more difficult to cover all points of space-time objects. Also, when we use more than 600 proposals, we encounter memory issues. In all of our experiments, 500 proposals are used due to the highest $\mathrm{S}_{assoc}$ score.

\begin{center}
\begin{table}[h]
\renewcommand{\arraystretch}{1.2}
\center
\normalsize{
\tabcolsep=0.10cm
 \begin{tabular}{c | c c c} 
 \toprule
 {Radius Size} &\textit{LSTQ} &  $\mathrm{S_{assoc}}$ & $\mathrm{S_{cls}}$ \\
 \midrule
{0.2$m$}  & {65.9} & {70.8} & {61.4} \\ 
{0.4$m$}  & {66.0} & {70.9} & {61.4} \\ 
\textbf{0.6$\boldsymbol{m}$}  & \textbf{66.4} & \textbf{71.8} & \textbf{61.4} \\ 
{0.8$m$}  & {66.3} & {71.6} & {61.4} \\ 
{1.0$m$}  & {66.1} & {71.3} & {61.3} \\ 

 \bottomrule
 \end{tabular}
\caption{\textbf{Ablation study on the impact of the radius size of proposals in \abbrev{}.}  In this ablation study, we aggregate 500 proposals based on the learned geometric features.}
\label{tab:ablate-radius-size}
}
\end{table}
\end{center}

\PAR{Effect of the radius size of proposals.} We examine the impact of the radius size during proposal generation in Tab.~\ref{tab:ablate-radius-size}. All the points voting for a space-time object center within the radius $r$ around the proposal center position are assigned to this proposal (see Sec.~\ref{sec:4D-StOP}). In Tab.~\ref{tab:ablate-radius-size}, we see that it is critical to select a suitable radius size during proposal generation. We achieve the best results with the radius size 0.6$m$. If we set the radius size under 0.6$m$, the $\mathrm{S}_{assoc}$ score decreases by approximately $1\%$ and $LSTQ$ score $0.5\%$. We believe that this decline results from having smaller proposals, \ie the proposals with a lower radius size, that cannot cover objects sufficiently.
The points that actually belong to a particular proposal are missed or assigned to another proposal, thus we fail to merge these proposals during the aggregation step resulting in object separations.
If we set the radius size too large, \eg $r=1.0m$, the points that actually belong to different objects are assigned to the same proposal, \ie one proposal covers multiple objects, thereby we predict several objects as one object. Also, we were inspired by the work in the 3D domain~\cite{Engelmann20203DMPAMA} which generates the proposals with a radius $r=0.3 m$. Compared to this work, our method obtains the best results with a radius twice as larger ($r=0.6 m$). It is anticipated to have a larger radius size as we work in a larger point cloud in the 4D domain. Overall, these ablations show that the radius size of proposals has a notable effect on the results and it should be adjusted considering the size of the 4D volume.

\PAR{Farthest Point Sampling (FPS) vs. Random Sampling.}
In order to generate proposals, we need to sample $K$ proposal centers from the center votes of all foreground points. We utilize the farthest point sampling (FPS) selection technique where the votes with the farthest point positions from each other are sampled. An alternative selection technique would be random sampling where we simply select a random set of $K$ votes. The results of both selection techniques are compared in Tab.~\ref{tab:ablate-selection}. The difference is negligibly small ($0.1\%$ $LSTQ$), showing the robustness of our method to the sampling strategies. Although we obtain similar results with both sampling strategies, we make use of the FPS selection technique as the probability of completely missing an object in a scene is much lower.

\begin{table}[ht!]
\renewcommand{\arraystretch}{1.2}
\center
\normalsize{
\tabcolsep=0.10cm
 \begin{tabular}{l | c c c} 
 \toprule
 {} & \textit{LSTQ} &  $\mathrm{S_{assoc}}$ & $\mathrm{S_{cls}}$ \\
 \midrule
 \textbf{Farthest-Point-Sampling} & \textbf{66.4} & \textbf{71.8} & \textbf{61.4} \\ 
 Random-Sampling & {66.3} & {71.5} & {61.4} \\
 \bottomrule
 \end{tabular}

\caption{\textbf{Ablation study on the impact of the proposal sampling strategy in \abbrev{}.} We aggregate $K=500$ proposals with $r=0.6 m$ based on the learned geometric features.}
\label{tab:ablate-selection}
}
\end{table}



\begin{figure}[ht!]
    \centering
    \includegraphics[width=1\linewidth]{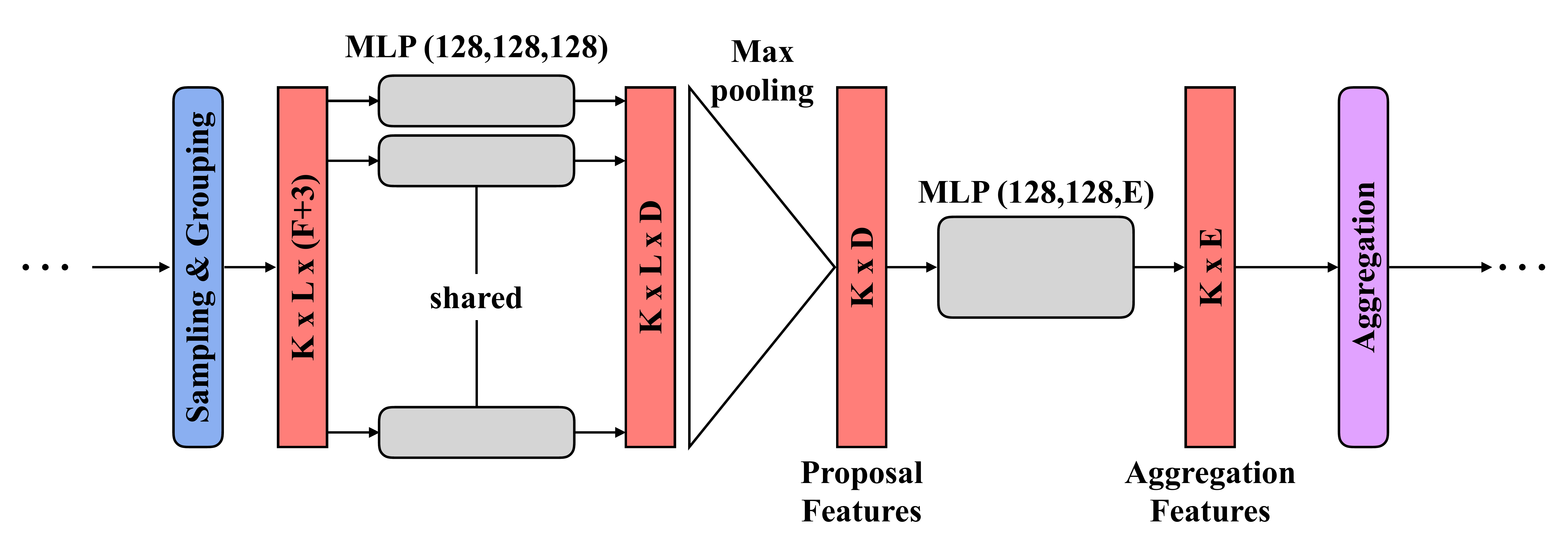}
    \caption{\textbf{The detailed architecture of Aggregation Feature Learning in \abbrev{}.}}
    \label{fig:4d-stop-agg-features}
\end{figure}


\section{Aggregation Feature Learning Details}
\label{sec:aggregation_network}
In Fig.~\ref{fig:4d-stop-agg-features}, we show the part of the \abbrev{} network in which the aggregation features are learned in detail. $K$ Proposals are aggregated based on learned aggregation features. Firstly, the features ($F=256$) and positions of the $L$ points associated with a proposal are passed to a shared MLP. The output features are further max pooled to get a single feature vector for each proposal which indicates the proposal features ($D=128$). These proposal features are passed to an additional MLP to get the final aggregation features ($E=7$).

\section{Per-Category Results}
\label{sec:per_category_results}
In Tab.~\ref{tab:per_category}, we provide the per-category results of \abbrev{} on the SemanticKITTI 4D Panoptic Segmentation test and validation sets. The scores are reported for 2 scan and 4 scan versions. The $\mathrm{S}_{assoc}$ scores are 0.00 for the objects from the \textit{stuff} classes as they are not evaluated for assigning associated instance IDs in time continuum.    


\begin{table}[htp]
\renewcommand{\arraystretch}{1.2}
\center
\footnotesize{
\tabcolsep=0.40cm
\begin{tabular}{l c |c c | c c} 
 \toprule
 {} & {} & \multicolumn{2}{c}{Test Set} & \multicolumn{2}{c}{Validation Set} \\
 {Category} & {\# Scans} &  $\mathrm{S_{assoc}}$ & $\mathrm{S_{cls}}$ & $\mathrm{S_{assoc}}$ & $\mathrm{S_{cls}}$ \\
 \midrule
 \multirow{2}{*}{Car} & {2} & {0.78} & {0.97} & {0.87} & {0.97}\\
 {} & {4} & {0.79} & {0.97} & {0.87} & {0.97}\\ \hline
 \multirow{2}{*}{Truck} & {2} & {0.48} & {0.53} & {0.10} & {0.32}\\
 {} & {4} & {0.56} & {0.46} & {0.17} & {0.31}\\ \hline
 \multirow{2}{*}{Bicycle} & {2} & {0.16} & {0.44} & {0.49} & {0.75}\\
 {} & {4} & {0.20} & {0.47} & {0.62} & {0.66}\\ \hline
 \multirow{2}{*}{Motorcycle} & {2} & {0.35} & {0.52} & {0.89} & {0.85}\\
 {} & {4} & {0.42} & {0.57} & {0.98} & {0.76}\\ \hline
 \multirow{2}{*}{Other-vehicle} & {2} & {0.35} & {0.55} & {0.47} & {0.66}\\
 {} & {4} & {0.42} & {0.53} & {0.56} & {0.63}\\ \hline
 \multirow{2}{*}{Person} & {2} & {0.34} & {0.57} & {0.39} & {0.70} \\
 {} & {4} & {0.40} & {0.55} & {0.47} & {0.66} \\ \hline
 \multirow{2}{*}{Bicyclist} & {2} & {0.62} & {0.59} & {0.80} & {0.89} \\ 
 {} & {4} & {0.66} & {0.62} & {0.75} & {0.88} \\ \hline
 \multirow{2}{*}{Motorcyclist} & {2} & {0.51} & {0.09} & {0.30} & {0.00}\\
 {} & {4} & {0.45} & {0.14} & {0.76} & {0.00} \\ \hline
 \multirow{2}{*}{Road} & {2} & {0.00} & {0.89} & {0.00} & {0.93} \\
 {} & {4} & {0.00} & {0.88} & {0.00} & {0.93}\\ \hline
 \multirow{2}{*}{Sidewalk} & {2} & {0.00} & {0.73} & {0.00} & {0.45} \\
 {} & {4} & {0.00} & {0.72} & {0.00} & {0.47}  \\ \hline
 \multirow{2}{*}{Parking} & {2} & {0.00} & {0.66} & {0.00} & {0.80}  \\
 {} & {4} & {0.00} & {0.66} & {0.00} & {0.80}  \\ \hline
 \multirow{2}{*}{Other-ground} & {2} & {0.00} & {0.31} & {0.00} & {0.01}  \\
 {} & {4} & {0.00} & {0.28} & {0.00} & {0.00}  \\ \hline
 \multirow{2}{*}{Building} & {2} & {0.00} & {0.92} & {0.00} & {0.90}  \\
 {} & {4} & {0.00} & {0.92} & {0.00} & {0.90}  \\ \hline
 \multirow{2}{*}{Vegetation} & {2} & {0.00} & {0.84} & {0.00} & {0.61}  \\
 {} & {4} & {0.00} & {0.84} & {0.00} & {0.60}  \\ \hline
 \multirow{2}{*}{Trunk} & {2} & {0.00} & {0.69} & {0.00} & {0.88} \\
 {} & {4} & {0.00} & {0.69} & {0.00} & {0.88}  \\ \hline
 \multirow{2}{*}{Terrain} & {2} & {0.00} & {0.68} & {0.00} & {0.69}  \\
 {} & {4} & {0.00} & {0.68} & {0.00} & {0.68}  \\ \hline
 \multirow{2}{*}{Fence} & {2} & {0.00} & {0.66} & {0.00} & {0.74}  \\
 {} & {4} & {0.00} & {0.68} & {0.00} & {0.76}  \\ \hline
 \multirow{2}{*}{Pole} & {2} & {0.00} & {0.57} & {0.00} & {0.64}  \\
 {} & {4} & {0.00} & {0.58} & {0.00} & {0.64} \\ \hline
 \multirow{2}{*}{Traffic-sign} & {2} & {0.00} & {0.55} & {0.00} & {0.49} \\
 {} & {4} & {0.00} & {0.54} & {0.00} & {0.51}  \\
 \bottomrule
\end{tabular}
\caption{\small\textbf{Per-category results of \abbrev{} on the SemanticKITTI test and validation set.}}
\label{tab:per_category}
}
\end{table}


\section{Additional Qualitative Results}
\label{sec:additional_qualitative}
\begin{figure}[ht!]
    \centering
    \includegraphics[width=1\linewidth]{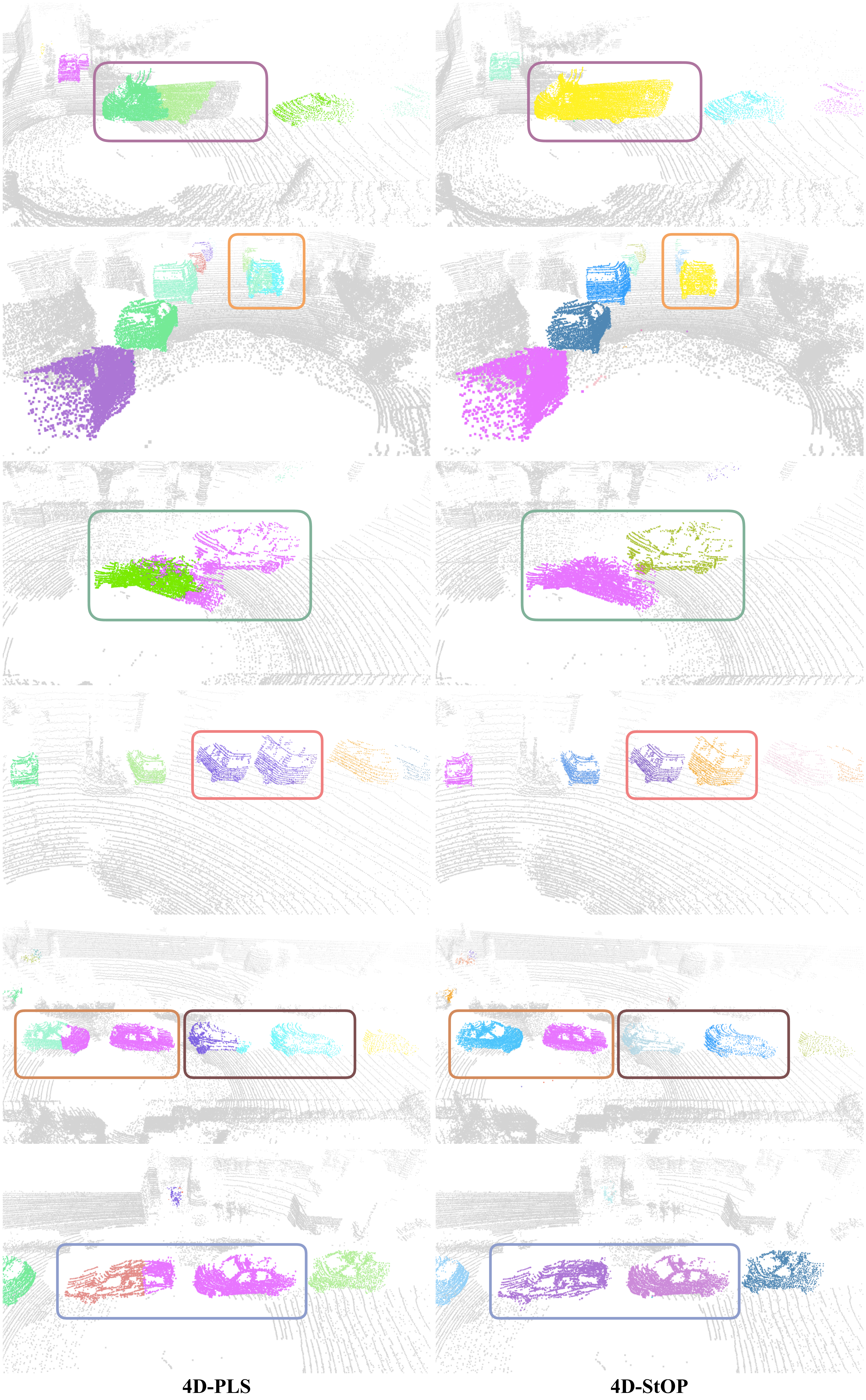}
    \caption{\textbf{Additional qualitative results on SemanticKITTI validation split.  } On the same scans, we compare the qualitative results for 4D-PLS (left) and \abbrev{} (right). We depict the particular cases within boxes. For simplicity, points with stuff class label are colored gray. All the scans are 4D volumes over 4 scans.}
    \label{fig:additional_qualitative}
\end{figure}

\begin{figure}[ht!]
    \centering
    \includegraphics[width=1\linewidth]{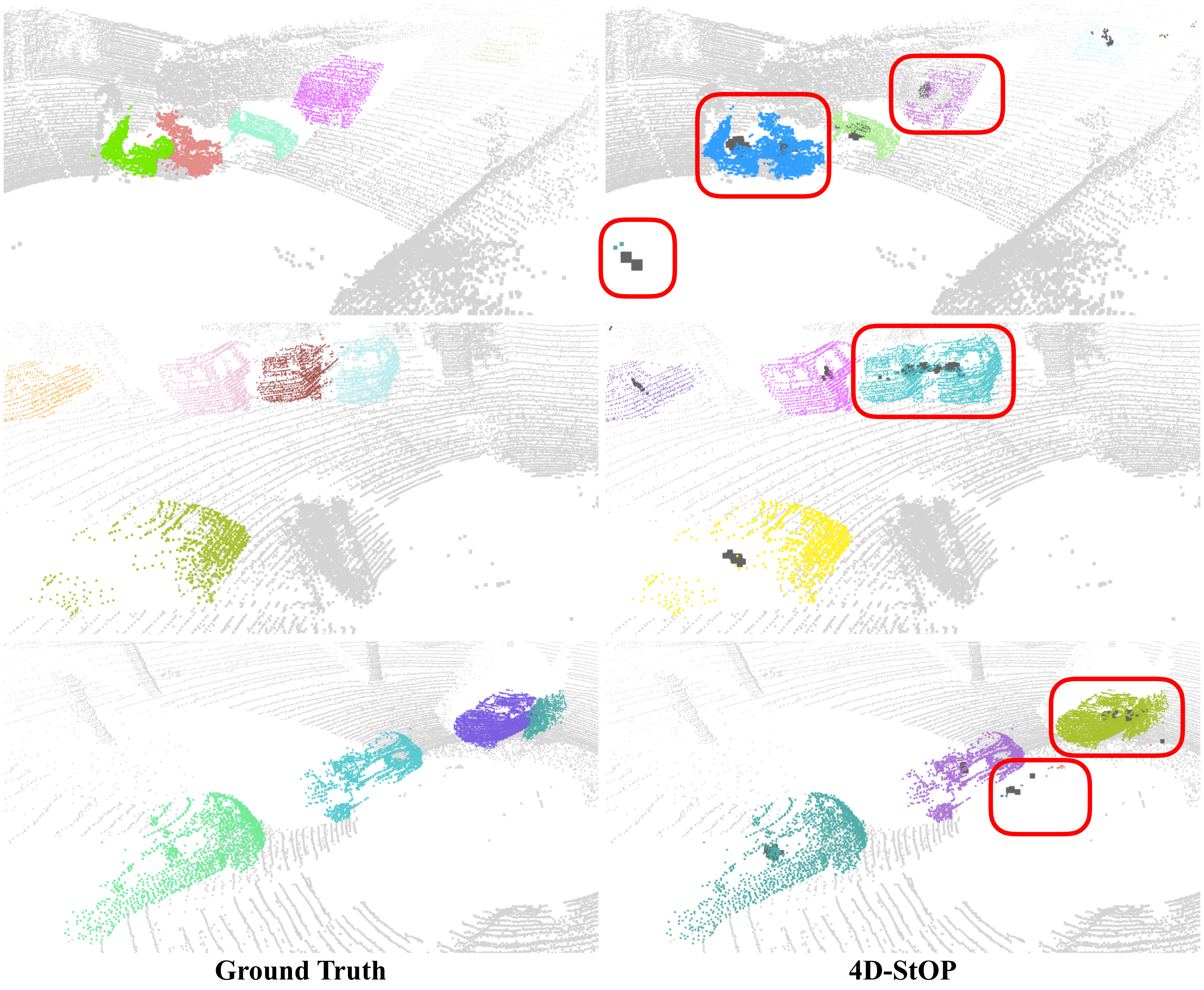}
    \vspace{-5pt}
    \caption{\textbf{Additional failure cases on SemanticKITTI validation split.} The scans with ground truths are shown on the left and the same scans with failure cases on the right. Each failure case is indicated within red boxes. The dark gray squares demonstrate the refined proposal centers. Again, the points with
    stuff class label are colored gray and all the scans are 4D volumes over 4 scans.
    }
    \label{fig:additional_failure}
\end{figure}

We depict the additional qualitative results on example scans for 4D-PLS and \abbrev{} in the first and second columns of Fig.~\ref{fig:additional_qualitative}. Each specific case is indicated in boxes. In Fig.~\ref{fig:additional_failure}, we also show the additional failure cases within red boxes. 

\end{document}